\documentclass[journal]{IEEEtran}
\usepackage{bbm}      
\usepackage{graphics} 
\usepackage{epsfig} 
\usepackage{amsmath} 
\usepackage{amssymb}  
\usepackage{amsfonts}
\usepackage{algorithm} 
\usepackage{algorithmic} 
\usepackage{subfigure}
\usepackage[left=0.75in,right=0.75in,top=1in,bottom=0.75in]{geometry}
\usepackage{multirow}
\usepackage{cite}

\ifCLASSINFOpdf
\else
\fi
\begin{document}
%
\title{Disturbance Rejection Control for Autonomous Trolley Collection Robots with Prescribed Performance}

%
%
%

\author {Rui-Dong~Xi$^{1}$,~\IEEEmembership{}
        Liang~Lu$^{1}$,~\IEEEmembership{}
        Xue~Zhang$^{1}$,~\IEEEmembership{Member,~IEEE,}
        Xiao~Xiao$^{1,3}$,~\IEEEmembership{Member,~IEEE,}
         Bingyi~Xia$^{1}$,~\IEEEmembership{}
        Jiankun~Wang$^{1,2}$,~\IEEEmembership{Senior Member,~IEEE,}
        and~Max Q.-H.~Meng$^{1}$,~\IEEEmembership{Fellow,~IEEE}
\thanks{This work was supported by Shenzhen Key Laboratory of Robotics Perception and Intelligence (ZDSYS20200810171800001), Southern University of Science and Technology, Shenzhen 518055, China.(\textit {Corresponding authors: Max Q.-H. Meng.})}
\thanks{Rui-Dong Xi, Liang Lu, Xue Zhang, Xiao Xiao, Jian-Kun Wang and Max Q.-H. Meng are with Shenzhen Key Laboratory of Robotics Perception and Intelligence, the Department of Electronic and Electrical Engineering, Southern University of Science and Technology, Shenzhen 518055, China. (e-mail: max.meng@ieee.org).}
\thanks{Jiankun Wang is also with Jiaxing Research Institute, Southern University of Science and Technology, Jiaxing, China}
\thanks{Xiao Xiao is also with Yuanhua Robotics, Perception $\&$ AI Technologies Ltd.  Shenzhen 518055, China.}
}

%
%

\markboth{ }%
{Shell \MakeLowercase{\textit{et al.}}: Bare Demo of IEEEtran.cls for IEEE Journals}
%



\maketitle

\begin{abstract}
Trajectory tracking control of autonomous trolley collection robots (ATCR) is an ambitious work due to the complex environment, serious noise and external disturbances.  This work investigates a control scheme for ATCR subjecting to severe environmental interference. A kinematics model based adaptive sliding mode disturbance observer with fast convergence is first proposed to estimate the lumped disturbances. On this basis, a robust controller with prescribed performance is proposed using a backstepping technique, which improves the transient performance and guarantees fast convergence. Simulation outcomes have been provided to illustrate the effectiveness of the proposed control scheme. 
\end{abstract}


%
\IEEEpeerreviewmaketitle

\section{Introduction}
%
%
%
%

\IEEEPARstart{T}{he} deployment of robot systems to achieve autonomous recycling luggage trolleys is of great significance for increasing reuse efficiency of luggage trolleys, lowering labor expenses, and building a smart airport \cite{wang2021real,xiao2022robotic}. However, this is an extremely difficult task since the extensive and intricate pedestrian flow poses great challenges to the perception, planning, and control of the autonomous trolley collection robots (ATCR)\cite{pan2020searching,wang2020coarse}. In the research work of \cite{xiao2022robotic}, an overall diagram of a differential steered ATCR is introduced and a model predictive controller (MPC) is developed as the preliminary control scheme, but the external disturbances and convergence rate are not fully considered. Hence, in this work, a disturbance rejection control scheme with prescribed performance is developed for robust and accurate control of the ATCR.

For an ATCR, there are some challenging problems that existed for example: i) the robot needs to travel effectively in crowded and cramped environments; ii) the robot needs to dispose of skidding and slipping effectively under different road conditions, like slippery ground. In addition, while transporting a chain of collected luggage trolleys, two ATCRs have to collaborate with each other. As a consequence, the control system is required to have fast response, high precision, and high interference cancellation abilities.

To date, various control methods have been adopted in mobile robots for the sake of robust and precise trajectory tracking \cite{2018Trajectory,huskic2019high,yue2022path}. In the reference \cite{2018Trajectory} and \cite{xi2019tracking},  second order sliding mode controllers are utilized, and in \cite{huskic2019high}, a robust control scheme is developed for skid-steered mobile robots on account of challenging terrains. These methods have improved the robot's resistance to skidding and slipping to a certain extent, and have compared different ground conditions. However, these commonly used methods are excessively dependent on the dynamics model and the accurate force between the robot and the ground. In practice, it is very difficult to accurately describe the dynamics model, and the force between the robot and the ground is constantly fluctuating with the change in the environment. Therefore, there exist significant model uncertainties and interference, and these factors have not been fully addressed. 

Disturbance observer (DOB) is an efficient method to tackle model uncertainties and disturbances \cite{chen2016disturbance,kang2014robust,lu2022robust,yu2018robust,Xi2023,xi2020adaptive}. In the literature \cite{chen2016disturbance}, Chen utilized a DOB to deal with the skidding, slipping, and input disturbances in a differential steering mobile robot. In \cite{kang2014robust} a fuzzy DOB is also utilized in the control of a differential steering mobile robot with skidding and slipping phenomenon. Nevertheless, the convergence rate has never been considered in these DOBs. Adaptive sliding mode disturbance observer (ASMDOB) as a new kind of DOB has been successfully developed to estimate lumped uncertainties in various electromechanical systems\cite{chen2015disturbance,liang2022adaptive,rabiee2019continuous}. For instance, in the research works \cite{zhu2018adaptive}, \cite{shi2022adaptive} and \cite{xi2022adaptive}, different ASMDOBs were introduced in the control of robot manipulators. In these works, in addition to the accurate observation of interference, fast convergence could also be guaranteed. In \cite{xi2022adaptive}, the backstepping technique is initially utilized in the design of the ASMDOB. In this work, except for the fast and accurate observation of the lumped disturbances, the tracking cost has also been reduced and the engineering applicability has been further improved.

In addition to interference cancellation, transient performance like overshoot and convergence rate should also be considered in the controller design of ATCRs. Exactly, the prescribed performance control (PPC) is an effective way to solve this problem \cite{bechlioulis2009adaptive,cheng2022prescribed,wang2021event}. For example, in \cite{wang2021event}, an event-based PPC is investigated for control of dynamic positioning vessels with unknown and time-varying sea loads.

Inspired by the aforementioned discussions, an ASMDOB based robust controller with prescribed performance is developed in this study. Instead of using the complicated dynamics model, the kinematics model is adopted which guarantees the simplicity and applicability of the algorithm. The advantages of ASMDOB, backstepping technique, and PPC method are integrated in this controller. The main contributions of this work can be summarized as follows: 
\begin{itemize}
    \item A kinematics model based ASMDOB with fast convergence is proposed, which has the advantage of higher engineering practicability.
    \item A robust controller is developed integrating the PPC, backstepping technique, and finite-time convergence method. The proposed control scheme can not only ensure the fast convergence of tracking errors, but also guarantee transient performance. 
\end{itemize}


\textit{Notation:} In this brief, $\left\| \cdot \right\|$ represents the Euclidean norm. The vectors ${\mathop{\rm sgn}} ( \sigma ) = {[{\rm{sign}}( \sigma_1 ), \cdots ,{\rm{sign}}( \sigma_n )]^{\rm{T}}}$, ${{\mathop{\rm sgn}} ^a}( \sigma ) = {[{\left|  \sigma_1  \right|^a}{\rm{sign}}( \sigma_1 ),  \cdots ,{\left|  \sigma_n  \right|^a}{\rm{sign}}( \sigma_n )]^{\rm{T}}}$.

\section{Problem Statement}
\label{sec_PS}



\begin{figure}[ht]
\centering
\includegraphics[width=3.4in]{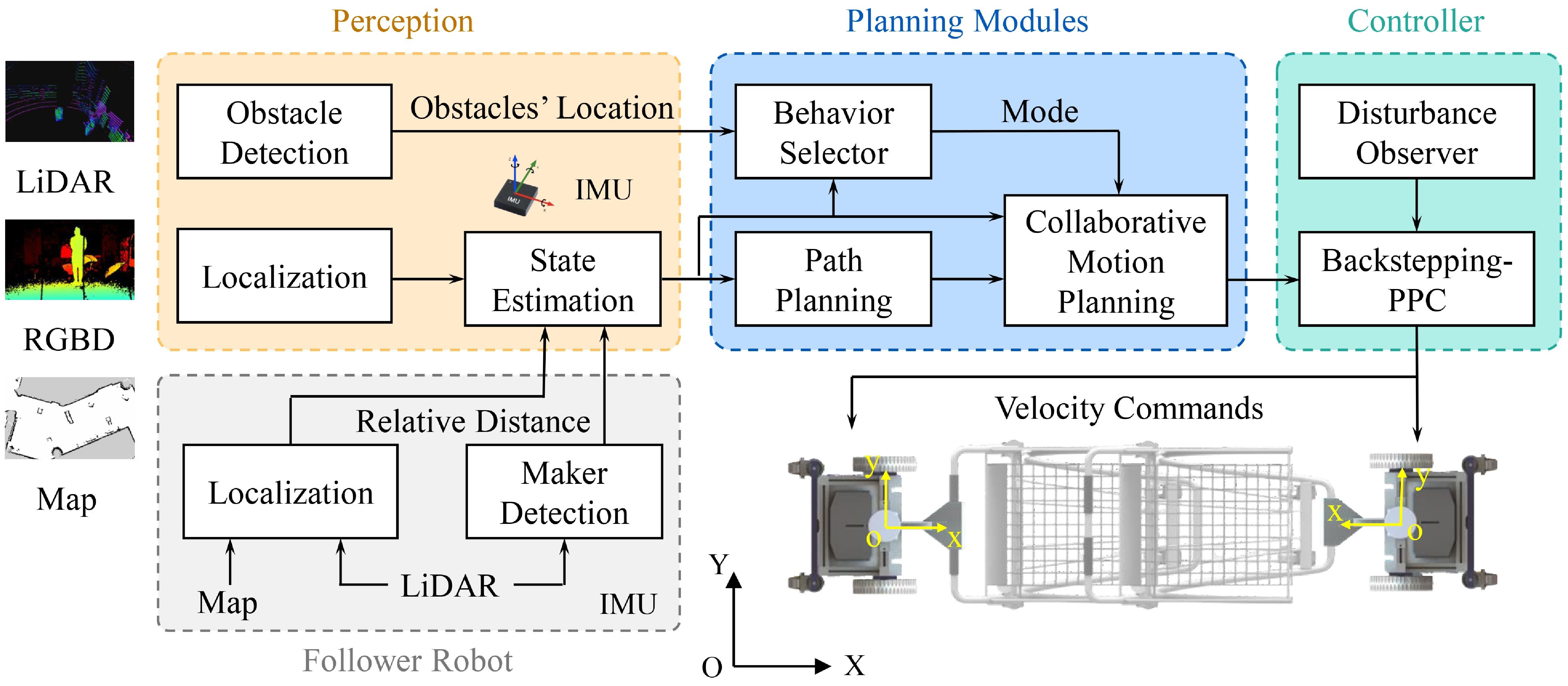}
\caption{Control system block diagram of the ATCR.}\label{mobile_rob}
\end{figure}

Overall control system block diagram of the developed ATCR is depicted in Fig.~\ref{mobile_rob}. This work mainly focus on the controller design. Define $XOY$ as the fixed frame and $xoy$ as a body frame, considering kinematics interference, kinematics model of the robot is given by
\begin{equation} \label{kine_model}
    \dot q = Tu + {d_v}
\end{equation}
where $q = {[\begin{array}{*{20}{c}}
   x & y & \theta   \\
\end{array}]^{\rm{T}}}$, $x$ and $y$ denote the center point position corresponding to the
fixed frame and $\theta$ is the robot's orientation, $u = {\left[ {\begin{array}{*{20}{c}}
   v & \omega   \\
\end{array}} \right]^{\rm{T}}}$ presents the linear and angular velocity of the robot, and $T = \left[ {\begin{array}{*{20}{c}}
   {\cos \theta } & 0  \\
   {\sin \theta } & 0  \\
   0 & 1  \\
\end{array}} \right]$. ${d_v} = \left[ {\begin{array}{*{20}{c}}
   { {d_{vx}}}  &
   { {d_{vy}}}  &
   {{d_\omega }}  \\
\end{array}} \right]^{\rm{T}}$ with $d_{vx}$, $d_{vy}$, $d_{\omega}$ present the velocity interference respectively. 

The main objective of this work is to design an active disturbance rejection control scheme for the ATCR with both the transient and steady state performances guaranteed. To proceed with the design of control scheme, the following assumption and lemmas are required:

\noindent {\textit{Assumption 1:}} The lumped interference $d_v$ and its derivative are unknown but bounded. 

\noindent {\textit{Lemma 1 (see \cite{xi2022adaptive}): }} For positive definite function $V(t)$ which fulfills:
\begin{equation}
\dot V(t) + \kappa_1 V(t) + \kappa_2 {V^\gamma }(t) \le 0,\forall t > {t_0}
\end{equation}
then $V(t)$ will converge to the equilibrium point in finite time $t_s$ with
\begin{equation}
{t_s} \le {t_0} + \frac{1}{{\kappa_1 (1 - \gamma )}}\ln \frac{{\kappa_1 {V^{1 - \gamma }}({t_0}) + \kappa_2 }}{\kappa_2 }
\end{equation}
where $\kappa_1$, $\kappa_2$, and $\gamma$ are design parameters with $\kappa_1 > 0$, $\kappa_2 > 0$, $0 < \gamma < 1$.

\noindent {\textit{Lemma 2 (see \cite{chen2016disturbance} and \cite{xi2022adaptive}):} } For the bounded initial conditions, if the Lyapunov function $V(x)$ satisfies ${V _1}(||x||) \le V(x) \le {V _2}(||x||)$ such that
\begin{equation}\label{eq5}
\dot V(x) \le  - {c_1}V(x) + {c_2}
\end{equation}
\noindent where ${V _1},{V _2}:{\Re^n} \to \Re$ are class $K$ functions and $c_1, c_2$ are positive constants, then the state $x(t)$ is uniformly bounded.

\section{Design of the observer based kinematics controller}
\label{sec_DC}

\subsection{Design of ASMDOB}
In this part, the ASMDOB is introduced to estimate the uncertain part $d_v$ in the kinematics model. At first, auxiliary variables $\sigma \in {\Re ^{3 \times 1}} $ and $z \in {\Re ^{3 \times 1}}$ are developed to reflect the system states as
\begin{equation}
    \sigma  = z - q.
\end{equation}
Considering the function candidate ${L_1} = {\textstyle{ \frac{1}{2} }}{\sigma ^{\rm{T}}}\sigma $ and derivative it with respect to time, one has
\begin{equation}\label{dot_L1}
    {{\dot L}_1} = {\sigma ^T}(\dot z - Tu - {d_v}).
\end{equation}

Define 
\begin{equation}\label{dot_z}
    \dot z = Tu - {c_1}\sigma  - {c_2}{{\mathop{\rm sgn}} ^{{\alpha _1}}}(\sigma ) - {k_d}{\mathop{\rm sgn}} (\sigma )
\end{equation}
where $c_1$, $c_2$ and $k_d$ are positive design parameters and $0<\alpha_1<1$. Substituting $\dot z$ into (\ref{dot_L1}) yields
\begin{equation}\label{dot_L1_2}
    {{\dot L}_1} =  - {c_1}{\sigma ^{\rm T}}\sigma  - {c_2}{\sigma ^{\rm T}}{{\mathop{\rm sgn}} ^{{\alpha _1}}}(\sigma ) - {\sigma ^{\rm T}}{d_v} - {k_d}\sigma ^{\rm T} {\mathop{\rm sgn}} (\sigma ).
\end{equation}
While $k_d$ is selected fulfills ${k_d} \ge \max \|{{\bar d}_{v}}\|$ (${{\bar d}_v} \in {\Re ^{3 \times 1}}$ denotes the maximum interference in each directions), it is easy to obtain
\begin{equation}
    {{\dot L}_1} \le  - {c_1}{\sigma ^{\rm T}}\sigma  - {c_2}{\sigma ^{\rm T}}{{\mathop{\rm sgn}} ^{{\alpha _1}}}(\sigma ) \le 0.
\end{equation}

Then we have
\begin{equation}
    {{\dot L}_1} + 2{c_{\min }}{L_1} + {2^{{\raise0.5ex\hbox{$\scriptstyle {(1 + {\alpha _1})}$}
\kern-0.1em/\kern-0.15em
\lower0.25ex\hbox{$\scriptstyle 2$}}}}{c_{\min }}L_1^{{\raise0.5ex\hbox{$\scriptstyle {(1 + {\alpha _1})}$}
\kern-0.1em/\kern-0.15em
\lower0.25ex\hbox{$\scriptstyle 2$}}} \le 0
\end{equation}
where ${c_{\min }} = \min \{ {c_1},{c_2}\}$. According to lemma 1, it can be concluded that $z$ will converge to $q$ in finite time. However, it can be seen in (\ref{dot_z}) that the variable structure term ${k_d}{\mathop{\rm sgn}} (\sigma )$ will greatly influence the convergent performance due to its discontinuous characteristics. As a result, a low pass filter is adopted as
\begin{equation}
    {\lambda _0}\dot \zeta {\rm{ + }}\zeta {\rm{ = }}\mu 
\end{equation}
where $\mu = - {c_1}\sigma  - {c_2}{{\mathop{\rm sgn}} ^{{\alpha _1}}}(\sigma ) - {k_d}{\mathop{\rm sgn}} (\sigma )$, $\lambda_0$ is the design parameter of the filter. 

Define $s = \dot \sigma  + {\lambda _1}\sigma $ and the ASMDOB is devised as
\begin{equation}\label{ASMDOB}
    {{\dot {\hat d}}_v} = {\lambda _2}(\zeta  + {\lambda _1}\sigma  - {{\hat d}_v}) - ({k_s} + \hat \beta ){\mathop{\rm sgn}} (s)
\end{equation}
where $\lambda_1$, $\lambda_2$ and $k_s$ are design parameters and $\hat \beta$ is the estimation of $\beta$ as the upper bound of $\dot {\hat d}_v$. From (\ref{ASMDOB}) we can obtain
\begin{equation}
    {{\dot {\tilde d}}_v} \dot =  - {\lambda _2}s - {\lambda _2}{{\tilde d}_v} - ({k_s} + \hat \beta ){\mathop{\rm sgn}} (s) + \dot d_v
\end{equation}
where $\tilde d_v = d_v - \hat d_v$.

\textit{Theorem 1:} For the kinematics model of the skid steered mobile robot (\ref{kine_model}), the ASMDOB is developed as in (\ref{ASMDOB}), while the observer gains are designed fulfills $\lambda_2 > 0$ and ${k_s} \ge {\lambda _2} \| {s - {{\tilde d}_v}} \|$, and the adaptive law is given by
\begin{equation}
    \dot {\hat \beta}  =  - {\lambda _3}\hat \beta  + \left\| s \right\|
\end{equation}
where $\lambda _3$ is a positive parameter. Then the estimation error of the uncertainty $\tilde d_v$ is uniformly bounded with exponentially convergent characteristics. 

\textit{Proof:} Considering the Lyapunov function
\begin{equation}
    {L_2} = \frac{1}{2}\tilde d_v^{\rm T}{\tilde d_v} + \frac{1}{2}{\tilde \beta ^2}
\end{equation}
where $\tilde \beta  = \beta  - \hat \beta $. The time derivative of $L_2$ is calculated as
\begin{equation}
\begin{aligned}
  {{\dot L}_2} &= \tilde d_v^{\rm T}{\left[ { - {\lambda _2}s - {\lambda _2}{{\tilde d}_v} - ({k_s} + \hat \beta ){\mathop{\rm sgn}} (s) + \dot d_v} \right]^{\rm T}} + \tilde \beta \dot {\tilde \beta}  \\ 
  &=  - 2{\lambda _2}\tilde d_v^{\rm T}{{\tilde d}_v} - {\lambda _2}\tilde d_v^{\rm T}(s - {{\tilde d}_v}) - {k_s}\tilde d_v^{\rm T}{\mathop{\rm sgn}} (s) \\
  &- \hat \beta \tilde d_v^{\rm T}{\mathop{\rm sgn}} (s) 
  + \tilde d_v^{\rm T}\dot d_v - \tilde \beta \dot {\hat \beta}  \\ 
\end{aligned}
\end{equation}

According to the convergent property of $\sigma$ and the equivalent output injection principle \cite{lu2009sliding}, it is observed that $s$ can be concesived as the equivalent value of $\tilde d_v$. When the value of $k_s$ is properly defined, we can obtain

\begin{equation}
    \begin{aligned}
 {{\dot L}_2} &\le  - 2{\lambda _2}\tilde d_v^{\rm T}{{\tilde d}_v} - \hat \beta \| {\tilde d_v^{\rm T}} \| + \tilde d_v^{\rm T}\beta  - \tilde \beta \dot {\hat \beta}  \\ 
  &\le  - 2{\lambda _2}\tilde d_v^{\rm T}{{\tilde d}_v} + \tilde \beta \| {\tilde d_v^{\rm T}} \| - \tilde \beta \dot {\hat \beta}  \\ 
  &\le  - 2{\lambda _2}\tilde d_v^{\rm T}{{\tilde d}_v} + {\lambda _3}\tilde \beta \hat \beta  \\ 
  &\le  - 2{\lambda _2}\tilde d_v^{\rm T}{{\tilde d}_v} - \frac{{{\lambda _3}}}{2}{{\tilde \beta }^2} + \frac{{{\lambda _3}}}{2}{\beta ^2} \\ 
  &\le  - \min \left\{ {4{\lambda _2},{\lambda _3}} \right\}(\frac{1}{2}\tilde d_v^{\rm T}{\tilde d_v} + \frac{1}{2}{\tilde \beta ^2}) + {\delta _0} \\ 
  &\le - \lambda_{min} L_2 + {\delta _0} \\
 \end{aligned}
\end{equation}
where ${\lambda _{\min }} = \min \left\{ {4{\lambda _2},{\lambda _3}} \right\}$, ${\delta _0} = \frac{{{\lambda _3}}}{2}{\beta ^2}$, and one can obtain $0 \le {L_2} \le \frac{{{\delta _0}}}{{{\lambda _{\min }}}} + \left[ {{L_2}(0) - \frac{{{\delta _0}}}{{{\lambda _{\min }}}}} \right]{{\rm{exp}}({ - {\lambda _{\min }}t})}$. According to lemma 2, it is proved that the approximation error is exponentially convergent.

\noindent {\textit{Remark 1:}} From Eq. (\ref{ASMDOB}) it can be noticed that the ASMDOB has the ability of filtering, so the high-frequency oscillation caused by $\rm{sgn(s)}$ can be eliminated by itself.


\subsection{Design of kinematic controller with prescribed performance}

Define $e = q - q_d$ as the trajectory tracking error, where $q_d = {[\begin{array}{*{20}{c}}
   x_d & y_d & \varphi   \\
\end{array}]^{\rm{T}}}$ presents the desired trajectory and $\varphi$ is an auxiliary variable which will be introduced in the following parts. According to the kinematics model, one has
\begin{equation}
    \dot e = Tu + d_v - \dot q_d.
\end{equation}

To improve transient response characteristics and steady-state tracking accuracy, a performance function $\rho(t)$ is introduced which makes the tracking error satisfying 
\begin{equation}
 - \epsilon_i \rho_i (t) < e_i(t) < \epsilon_i \rho_i (t)
\end{equation}
where $e_i, (i =1,2,3)$ denotes $i$th element in $e$, $0 < \epsilon_i  \le 1$ is a design constant, $\rho_i (t)$ is chosen as $\rho_i (t) = ({\rho_{i0}} - {\rho _{i\infty} })\exp ( - {k_\rho }t) + {\rho _\infty }$ with $\rho_{i0} > \rho _{i\infty} >0 $ and $k_\rho > 0$.

Then define 
\begin{equation}
    {\eta _i} = \frac{1}{2}\ln \frac{{{\varepsilon _i}{\rho _i}(t) + e_i}}{{{\varepsilon _i}{\rho _i}(t) - e_i}}
\end{equation}
as the $i$th element of transformed error $\eta$. The dynamics of the transformed error can be obtained as
\begin{equation}\label{eq22}
    \dot \eta  = \Phi  + \Lambda (Tu + {d_v} - {{\dot q}_d})
\end{equation}
where $\Phi  = {\left[ {\begin{array}{*{20}{c}}
   {{\phi _1}} & {{\phi _2}} & {{\phi _3}}  \\
\end{array}} \right]^{\rm{T}}}$ with ${\phi _i} =  - \frac{{{\varepsilon _i}{{\dot \rho }_i}(t){e_i}}}{{\varepsilon _i^2\rho _i^2(t) - e_i^2}}$, and $\Lambda  = {\rm{diag}}\left\{ {\begin{array}{*{20}{c}}
   {{\Lambda _1}} & {{\Lambda _2}} & {{\Lambda _3}}  \\
\end{array}} \right\}$ with ${\Lambda _i} = \frac{{{\varepsilon _i}}}{{\varepsilon _i^2\rho _i^2(t) - e_i^2}}$.

According to Eq. (\ref{kine_model}), the desired trajectory can be obtained as
\begin{equation}
    \left\{ \begin{aligned}
 {{\dot x}_d} &= v\cos {\theta _d} \\ 
 {{\dot y}_d} &= v\sin {\theta _d} \\ 
 \end{aligned} \right.
\end{equation}
As coordinates ${x_d}$, ${y_d}$ and the desired direction angle $\theta_d$ are not independent with each other,  $({x_d},{y_d})$ is chosen as the target instruction. Then the controller is derived using backstepping technique as follows.

Step 1: Introducing the auxiliary control variable $\varphi$, and according to the kinematics model (\ref{kine_model}), define
\begin{equation}
    \left\{ \begin{aligned}
 \dot x &= v\cos \varphi  + {d_{v1}} \\ 
 \dot y &= v\sin \varphi  + {d_{v2}} \\ 
 \end{aligned} \right.
\end{equation}
where $d_{vi},(i = 1, 2, 3)$ denotes $i$th element in $d_v$.  Define a Lyapunov function candidate as
\begin{equation}
    {V_1} = \frac{1}{2}\eta _1^2 + \frac{1}{2}\eta _2^2.
\end{equation}
Differentiating $V_1$ with respect to time, we have
\begin{equation}\label{eq.26}
    \begin{aligned}
 {{\dot V}_1} &= {\eta _1}\left[ {{\phi _1} + {\Lambda _1}(v\cos \varphi  + {d_{v1}} - {{\dot x}_d})} \right] \\ 
  &+ {\eta _2}\left[ {{\phi _2} + {\Lambda _2}(v\sin \varphi  + {d_{v2}} - {{\dot y}_d})} \right]. \\ 
 \end{aligned}
\end{equation}
Define
\begin{subequations}\label{eq.27}
\begin{equation}
\begin{aligned}
 v\cos \varphi  &=  {{\dot x}_d}  + \Lambda _1^{ - 1}[ - {\phi _1} - {k_1}{\eta _1} - {k_2}{{\mathop{\rm sgn}} ^p}({\eta _1})  \\ 
  &- {k_3}{\mathop{\rm sgn}} ({\eta _1})] - {{\hat d}_{v1}} \\
\end{aligned}
\end{equation}
\begin{equation}
\begin{aligned}
 v\sin \varphi  &=  {{\dot y}_d}  + \Lambda _2^{ - 1}[ - {\phi _2} - {k_1}{\eta _2} - {k_2}{{\mathop{\rm sgn}} ^p}({\eta _2})  \\ 
  &- {k_3}{\mathop{\rm sgn}} ({\eta _2})] - {{\hat d}_{v2}} \\
\end{aligned}
\end{equation}
\end{subequations}
where $k_1$, $k_2$ and $k_3$ are design parameters, $\hat d_{vi},(i = 1, 2, 3)$ is the $i$th element in $\hat d_v$ and $0<p<1$. Substituting (\ref{eq.27}) into Eq. (\ref{eq.26}), one can obtain
\begin{equation}\label{eq.28}
    \begin{aligned}
 {{\dot V}_1} 
  &= {\eta _1}[ - {k_1}{\eta _1} - {k_2}{{\mathop{\rm sgn}} ^p}({\eta _1}) - {k_3}{\mathop{\rm sgn}} ({\eta _1}) + {{\tilde d}_{v1}}] \\ 
  &+ {\eta _2}[ - {k_1}{\eta _2} - {k_2}{{\mathop{\rm sgn}} ^p}({\eta _2}) - {k_3}{\mathop{\rm sgn}} ({\eta _2}) + {{\tilde d}_{v2}}] \\ 
  &=  - {k_1}\eta _1^2 - {k_2}{\eta _1}{{\mathop{\rm sgn}} ^p}({\eta _1}) - {k_3}{\eta _1}{\mathop{\rm sgn}} ({\eta _1}) + {\eta _1}{{\tilde d}_{v1}} \\ 
  &- {k_1}\eta _2^2 - {k_2}{\eta _2}{{\mathop{\rm sgn}} ^p}({\eta _2}) - {k_3}{\eta _2}{\mathop{\rm sgn}} ({\eta _2}) + {\eta _1}{{\tilde d}_{v2}}. \\ 
 \end{aligned}
\end{equation}
It can be noticed that while the value of $k_3$ is designed larger than ${\rm max}\{{\tilde d}_{v1}, {\tilde d}_{v2}\}$, one can obtain
\begin{equation}
    \begin{aligned}
 {{\dot V}_1} &\le  - {k_1}\eta _1^2 - {k_2}{\left| {{\eta _1}} \right|^{1 + p}} - {k_1}\eta _2^2 - {k_2}{\left| {{\eta _2}} \right|^{1 + p}} \\ 
  &\le  - 2{k_1}{V_1} - 2{k_2}V_1^{{\raise0.5ex\hbox{$\scriptstyle {(1 + p)}$}
\kern-0.1em/\kern-0.15em
\lower0.25ex\hbox{$\scriptstyle 2$}}}. \\ 
 \end{aligned}
\end{equation}

According to lemma 1, it can be concluded that $\eta_1$ and $\eta_2$ will converge to the equilibrium point in finite time. Let
\begin{subequations}\label{eq.30}
\begin{equation}
\begin{aligned}
 m_1  &=  {{\dot x}_d}  + \Lambda _1^{ - 1}[ - {\phi _1} - {k_1}{\eta _1} - {k_2}{{\mathop{\rm sgn}} ^p}({\eta _1})  \\ 
  &- {k_3}{\mathop{\rm sgn}} ({\eta _1})] - {{\hat d}_{v1}} \\
\end{aligned}
\end{equation}
\begin{equation}
\begin{aligned}
 m_2  &=  {{\dot y}_d}  + \Lambda _2^{ - 1}[ - {\phi _2} - {k_1}{\eta _2} - {k_2}{{\mathop{\rm sgn}} ^p}({\eta _2})  \\ 
  &- {k_3}{\mathop{\rm sgn}} ({\eta _2})] - {{\hat d}_{v2}}. \\
\end{aligned}
\end{equation}
\end{subequations}
While the linear velocity and virtual control law is given by
\begin{subequations}\label{eq.31}
\begin{equation}
v = \sqrt {m_1^2 + m_2^2} 
\end{equation}
\begin{equation}\label{eq.31b}
\varphi  = \arctan \frac{{{m_2}}}{{{m_1}}}
\end{equation}
\end{subequations}
Eq. (\ref{eq.27}) can be guaranteed. From Eq. (\ref{eq.31b}), it can be seen that while $\varepsilon _1$, $\rho_1$ are selected equal to $\varepsilon _2$, $\rho_2$, and the lumped disturbances are well observed, $\varphi$ will converge to $\theta_d$. To make $\theta$ track $\varphi$ fast and effectively, the following step is proposed.

Step 2: Define the Lyapunov function as
\begin{equation}\label{eq.32}
    {V_2} = {V_1} + \frac{1}{2}\eta _3^2.
\end{equation}
Differentiating $V_2$ with respect to time and considering (\ref{eq.28}), one has
\begin{equation}\label{eq.33}
\begin{aligned}
    {{\dot V}_2} &\le - {k_1}\eta _1^2 - {k_2}{\eta _1}{{\mathop{\rm sgn}} ^p}({\eta _1})  - {k_1}\eta _2^2 - {k_2}{\eta _2}{{\mathop{\rm sgn}} ^p}({\eta _2}) \\  
    &+ {\eta _3}[{\phi _3} + {\Lambda _3}(\omega  + {d_{v3}} - \dot \varphi )].
\end{aligned}
\end{equation}
Design the angular velocity control law as
\begin{equation}\label{eq.34}
\begin{aligned}
 \omega  &=  {{\dot \varphi}} - {{\hat d}_{v3}} + \Lambda _3^{ - 1}[ - {\phi _3} - {k_1}{\eta _3} - {k_2}{{\mathop{\rm sgn}} ^p}({\eta _3})  \\ 
  &- {k'_3}{\mathop{\rm sgn}} ({\eta _3})].  \\
\end{aligned} 
\end{equation}
Substituting (\ref{eq.34}) into Eq. (\ref{eq.33}) and select $k'_3$ larger than $\tilde d_{v3}$, one can obtain
\begin{equation}
    \begin{aligned}
 {{\dot V}_2} &\le  - {k_1}\eta _1^2 - {k_2}{\left| {{\eta _1}} \right|^{1 + p}} - {k_1}\eta _2^2 - {k_2}{\left| {{\eta _2}} \right|^{1 + p}} \\ 
 &- {k_1}\eta _3^2 - {k_2}{\left| {{\eta _3}} \right|^{1 + p}} \le  - 2{k_1}{V_2} - 2{k_2}V_2^{{\raise0.5ex\hbox{$\scriptstyle {(1 + p)}$}
\kern-0.1em/\kern-0.15em
\lower0.25ex\hbox{$\scriptstyle 2$}}}. \\ 
 \end{aligned}
\end{equation}
Based on lemma 1, we can conclude that both $\eta_1$, $\eta_2$ and $\eta_3$ will converge to the equilibrium point in finite time. Then the convergence of the whole system can be guaranteed. 





\section{Simulation Results}
\label{sec_SR}

\begin{figure}[ht]
\centering
\includegraphics[width=2.8in]{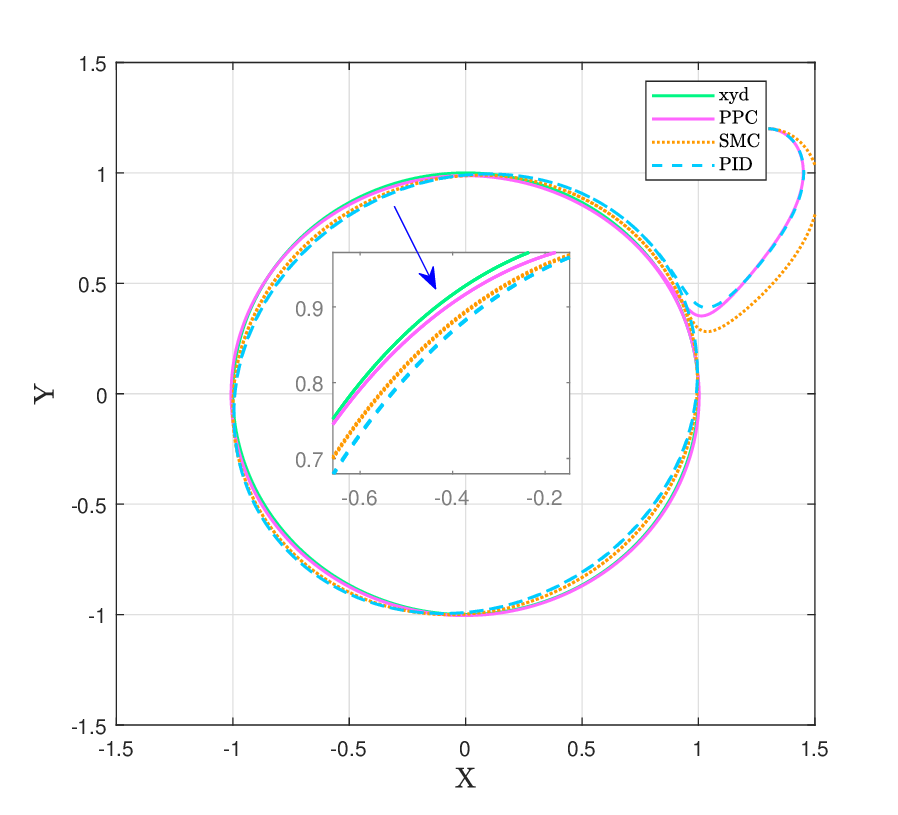}
\caption{Tracking results of the circular trajectory.}\label{Circle}
\end{figure}

\begin{figure}[!ht]
\centering
\subfigure[Observation results of $d_{v1}$]{
\includegraphics[width=2.8in]{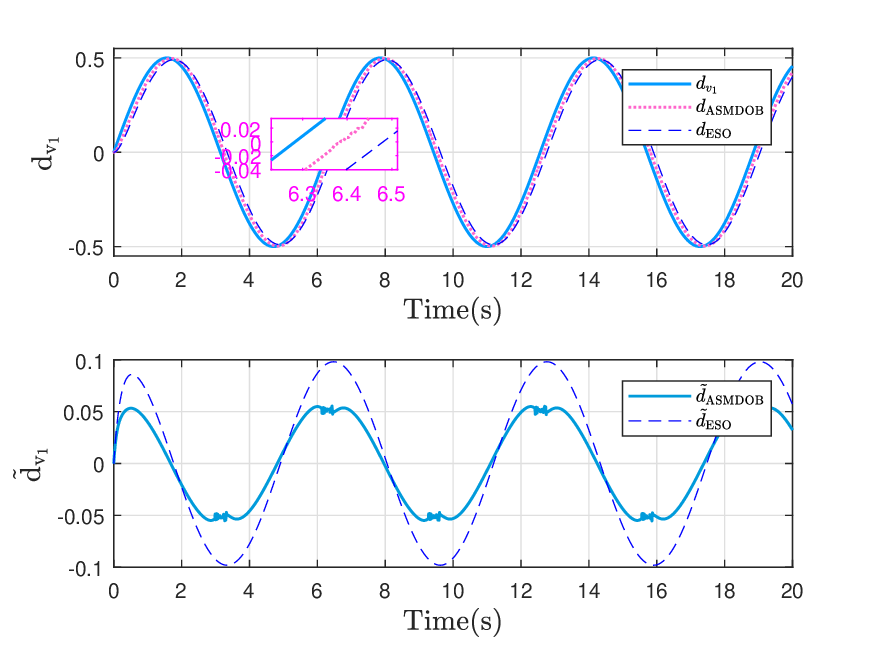}}
\subfigure[Observation results of $d_{v2}$]{
\includegraphics[width=2.8in]{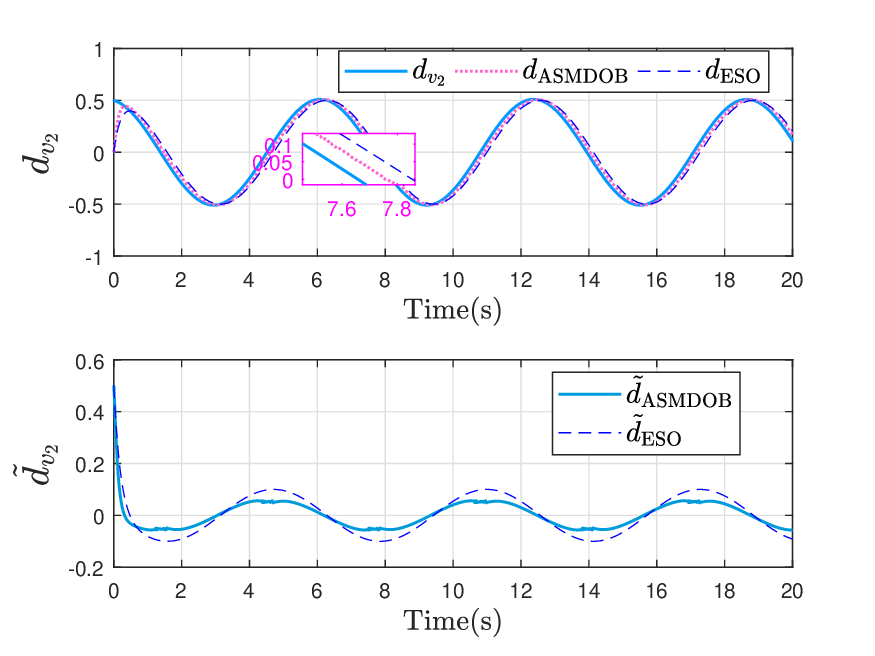}}
\subfigure[Observation results of $d_{v3}$]{
\includegraphics[width=2.8in]{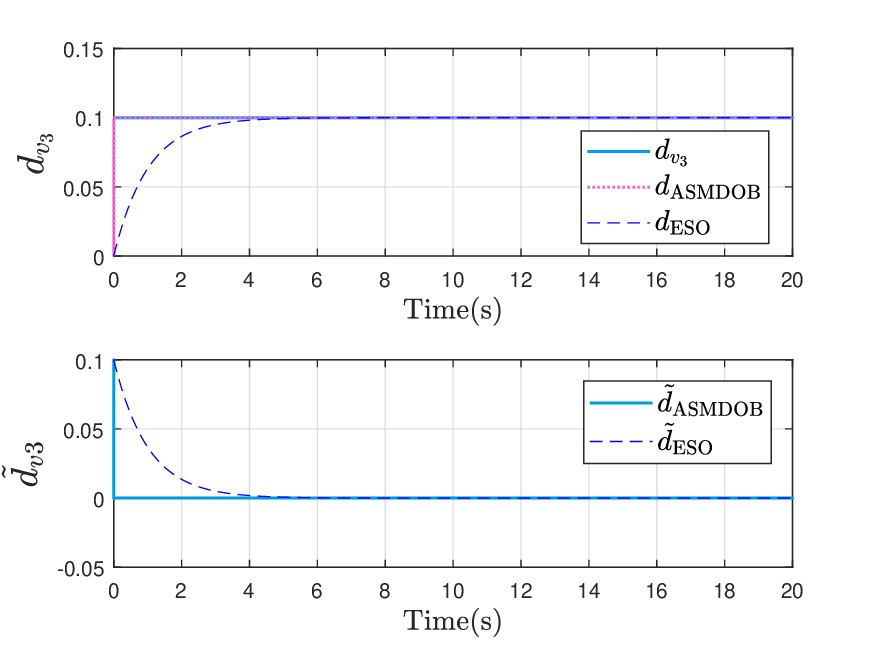}}
\caption{Observation results of external disturbances.}\label{Disturbance_observe}
\end{figure}

To validate the efficiency of the proposed control scheme, simulation comparisons with the mostly used PID and SMC controllers have been conducted. A circular shape target trajectory is selected as
\begin{equation}
\begin{aligned}
    x_r = \cos (t), 
    y_r = \sin (t). \\
\end{aligned}
\end{equation}
External disturbances are given by
\begin{equation}
    \left\{ \begin{aligned}
 {d_{v1}} &= 0.5\sin (t) \\ 
 {d_{v2}} &= 0.5\cos (t) + 0.1\cos (t + \frac{\pi }{2}) \\ 
 {d_{v3}} &= 0.1 \\ 
 \end{aligned} \right.
\end{equation}
It can be seen that both periodic and constant interference with high amplitude are considered. Tracking results of the circular trajectory is illustrated in Fig,(\ref{Circle}). To demonstrate the superiority of the developed DOB, the commonly used extended disturbance observer (ESO) is utilized to make a comparison. Observation results based on the proposed ASMDOB and ESO are demonstrated in the Fig. (\ref{Disturbance_observe}). It can be seen in these figures that all the disturbances in the $x$, $y$ and $\theta$ directions are well estimated. Moreover, from the comparison results it can be noticed that the proposed ASMDOB has faster convergence speed and higher estimation accuracy. 

\begin{figure}[ht]
\centering
\subfigure[Tracking result in $x$ and $y$ axis]{
\includegraphics[width=2.8in]{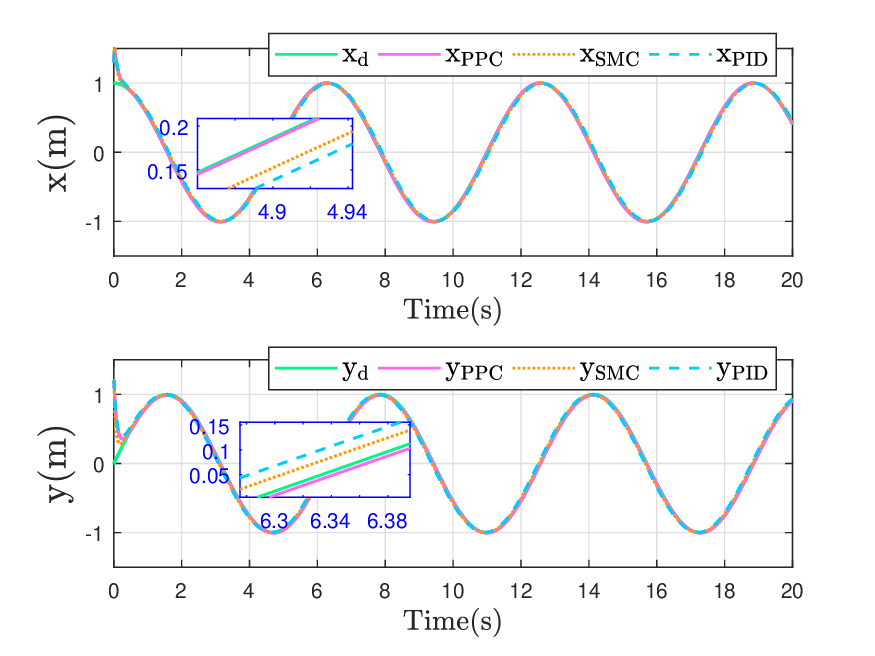} \label{xyaxis}}
\subfigure[Linear and angular velocities]{
\includegraphics[width=2.8in]{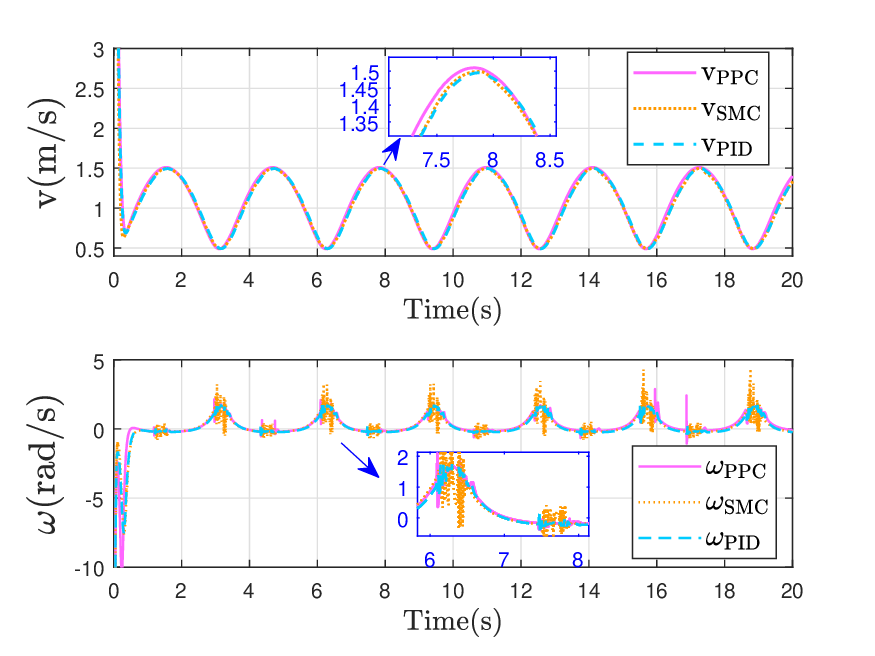}\label{velocity}}
\caption{Tracking results of the three controllers.}\label{Result}
\end{figure}

\begin{figure}[ht]
\centering
\includegraphics[width=2.8in]{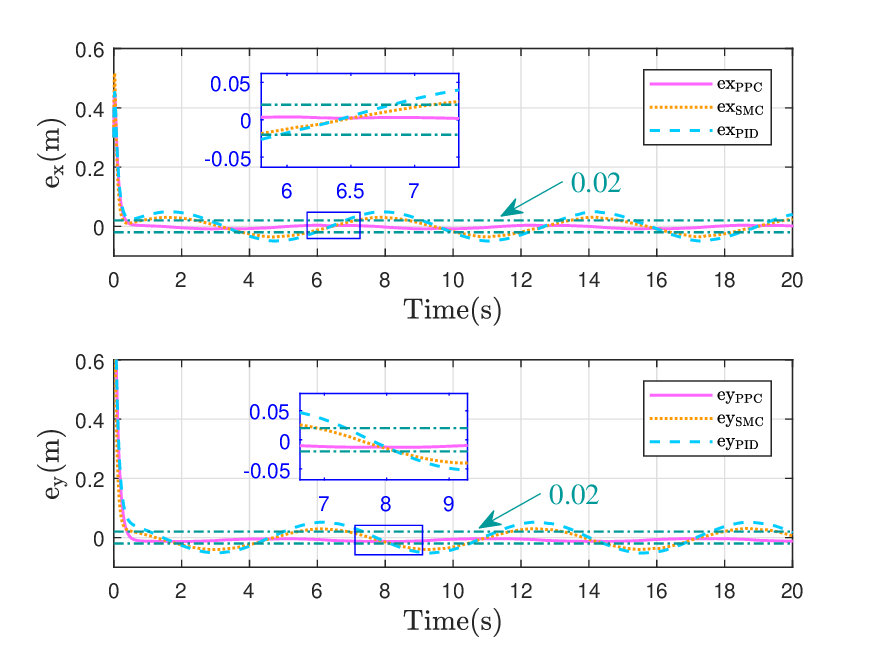}
\caption{Tracking errors of the three controllers.}\label{Error}
\end{figure}

\begin{figure}[ht]
\centering
\includegraphics[width=2.8in]{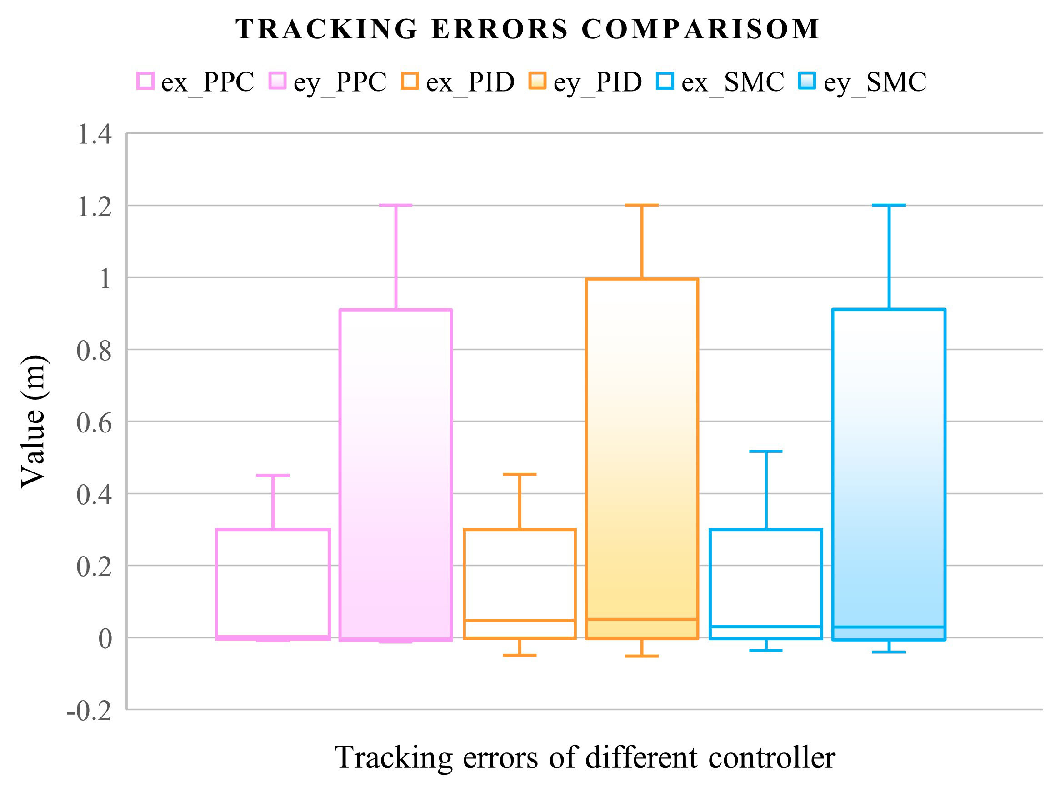}
\caption{Tracking error box-plots of the three controllers.}\label{Boxplots}
\end{figure}

The tracking processes in $x$ and $y$ orientation as well as the linear and angular velocity are provided in Fig.(\ref{Result}). The starting point is set at ($1.3$, $1.2$).  To ensure the reliability of the comparison, the parameters of the controllers have been adjusted to their best performance. From these figures, it can be found that although the objective trajectory can be tracked with all these methods, the tracking performance is totally different. The tracking accuracy of the proposed PPC is obviously higher and the input is smoother than the traditional SMC. To provide a clearer comparison result, the tracking errors are shown in the Fig. (\ref{Error}), and the root mean square error (RMS) value, maximum (MAX) value, and the mean value of the errors are given in the Table \ref{tab:1} (unit: $m$). Exactly, while the system converges, the tracking accuracy of PPC is kept within 0.01 $\rm{m}$, which is much lower than that of similar controllers, and the box-plots of the tracking errors is given in Fig.(\ref{Boxplots}). Therefore, it can be concluded that the proposed PPC could fully meet the control target of the ATCR.

\begin{table}[]
\caption{Comparison of the tracking errors.}\label{tab:1}
\renewcommand{\arraystretch}{1.2}
\begin{tabular}{c|ccc|ccc}
\hline
\multirow{2}{*}{Method} & \multicolumn{3}{c|}{$x$-axis}                                        & \multicolumn{3}{c}{$y$-axis}                                        \\ \cline{2-7} 
                            & \multicolumn{1}{c|}{RMS}    & \multicolumn{1}{c|}{MAX}    & MEAN   & \multicolumn{1}{c|}{RMS}    & \multicolumn{1}{c|}{MAX}   & MEAN   \\ \hline
PPC                         & \multicolumn{1}{c|}{0.2236} & \multicolumn{1}{c|}{0.4505} & 0.1371 & \multicolumn{1}{c|}{0.6338} & \multicolumn{1}{c|}{1.200} & 0.3812 \\
SMC                         & \multicolumn{1}{c|}{0.2520} & \multicolumn{1}{c|}{0.5170} & 0.1612 & \multicolumn{1}{c|}{0.6368} & \multicolumn{1}{c|}{1.200} & 0.3953 \\
PID                         & \multicolumn{1}{c|}{0.2383} & \multicolumn{1}{c|}{0.4522} & 0.1552 & \multicolumn{1}{c|}{0.6683} & \multicolumn{1}{c|}{1.200} & 0.4285 \\ \hline
\end{tabular}
\end{table}

\section{Conclusion and Future works}
\label{sec_Con}
This paper investigates an adaptive sliding mode disturbance observer based robust controller with prescribed performance for autonomous trolley collection mobile robots. To improve the engineering practicability, kinematics model is utilized and the disturbance observer is developed to guarantee the robustness. Prescribed performance control method integrated with backstepping technique is adopted to generate the control input, and in such manner both the stable and transient performance are guaranteed. In the future work, experiments will be conducted on the real robot platform to test the reliability of the proposed control scheme.


%





\ifCLASSOPTIONcaptionsoff
  \newpage
\fi



%




\bibliographystyle{IEEEtran}

\bibliography{DRCT}

\begin{thebibliography}{10}
\providecommand{\url}[1]{#1}
\csname url@samestyle\endcsname
\providecommand{\newblock}{\relax}
\providecommand{\bibinfo}[2]{#2}
\providecommand{\BIBentrySTDinterwordspacing}{\spaceskip=0pt\relax}
\providecommand{\BIBentryALTinterwordstretchfactor}{4}
\providecommand{\BIBentryALTinterwordspacing}{\spaceskip=\fontdimen2\font plus
\BIBentryALTinterwordstretchfactor\fontdimen3\font minus
  \fontdimen4\font\relax}
\providecommand{\BIBforeignlanguage}[2]{{%
\expandafter\ifx\csname l@#1\endcsname\relax
\typeout{** WARNING: IEEEtran.bst: No hyphenation pattern has been}%
\typeout{** loaded for the language `#1'. Using the pattern for}%
\typeout{** the default language instead.}%
\else
\language=\csname l@#1\endcsname
\fi
#2}}
\providecommand{\BIBdecl}{\relax}
\BIBdecl

\bibitem{wang2021real}
J.~Wang and M.~Q.-H. Meng, ``Real-time decision making and path planning for
  robotic autonomous luggage trolley collection at airports,'' \emph{IEEE
  Transactions on Systems, Man, and Cybernetics: Systems}, vol.~52, no.~4, pp.
  2174--2183, 2021.

\bibitem{xiao2022robotic}
A.~Xiao, H.~Luan, Z.~Zhao, Y.~Hong, J.~Zhao, W.~Chen, J.~Wang, and M.~Q.-H.
  Meng, ``Robotic autonomous trolley collection with progressive perception and
  nonlinear model predictive control,'' in \emph{2022 International Conference
  on Robotics and Automation (ICRA)}.\hskip 1em plus 0.5em minus 0.4em\relax
  IEEE, 2022, pp. 4480--4486.

\bibitem{pan2020searching}
J.~Pan, X.~Mai, C.~Wang, Z.~Min, J.~Wang, H.~Cheng, T.~Li, E.~Lyu, L.~Liu, and
  M.~Q.-H. Meng, ``A searching space constrained partial to full registration
  approach with applications in airport trolley deployment robot,'' \emph{IEEE
  Sensors Journal}, vol.~21, no.~10, pp. 11\,946--11\,960, 2020.

\bibitem{wang2020coarse}
C.~Wang, X.~Mai, D.~Ho, T.~Liu, C.~Li, J.~Pan, and M.~Q.-H. Meng,
  ``Coarse-to-fine visual object catching strategy applied in autonomous
  airport baggage trolley collection,'' \emph{IEEE Sensors Journal}, vol.~21,
  no.~10, pp. 11\,844--11\,857, 2020.

\bibitem{2018Trajectory}
I.~Matraji, A.~Al-Durra, A.~Haryono, K.~Al-Wahedi, and M.~Abou-Khousa,
  ``Trajectory tracking control of skid-steered mobile robot based on adaptive
  second order sliding mode control,'' \emph{Control Engineering Practice},
  vol.~72, no. MAR., pp. 167--176, 2018.

\bibitem{huskic2019high}
G.~Huski{\'c}, S.~Buck, M.~Herrb, S.~Lacroix, and A.~Zell, ``High-speed path
  following control of skid-steered vehicles,'' \emph{The International Journal
  of Robotics Research}, vol.~38, no.~9, pp. 1124--1148, 2019.

\bibitem{yue2022path}
X.~Yue, J.~Chen, Y.~Li, R.~Zou, Z.~Sun, X.~Cao, and S.~Zhang, ``Path tracking
  control of skid-steered mobile robot on the slope based on fuzzy system and
  model predictive control,'' \emph{International Journal of Control,
  Automation and Systems}, vol.~20, no.~4, pp. 1365--1376, 2022.

\bibitem{xi2019tracking}
R.~Xi and L.~Tang, ``Tracking control of a skid steered mobile robot with
  adaptive robust second order sliding-mode controller,'' in \emph{2019 IEEE
  International Conference on Industrial Engineering and Engineering Management
  (IEEM)}.\hskip 1em plus 0.5em minus 0.4em\relax IEEE, 2019, pp. 293--297.

\bibitem{chen2016disturbance}
M.~Chen, ``Disturbance attenuation tracking control for wheeled mobile robots
  with skidding and slipping,'' \emph{IEEE Transactions on Industrial
  Electronics}, vol.~64, no.~4, pp. 3359--3368, 2016.

\bibitem{kang2014robust}
H.-S. Kang, C.-H. Hyun, and S.~Kim, ``Robust tracking control using fuzzy
  disturbance observer for wheeled mobile robots with skidding and slipping,''
  \emph{International Journal of Advanced Robotic Systems}, vol.~11, no.~5,
  p.~75, 2014.

\bibitem{lu2022robust}
J.~Lu, Y.~Liu, W.~Huang, K.~Bi, Y.~Zhu, and Q.~Fan, ``Robust control strategy
  of gradient magnetic drive for microrobots based on extended state
  observer,'' \emph{Cyborg and Bionic Systems}, 2022.

\bibitem{yu2018robust}
L.~Yu, J.~Huang, and S.~Fei, ``Robust switching control of the direct-drive
  servo control systems based on disturbance observer for switching gain
  reduction,'' \emph{IEEE Transactions on Circuits and Systems II: Express
  Briefs}, vol.~66, no.~8, pp. 1366--1370, 2018.

\bibitem{Xi2023}
R.-D. Xi, T.-N. Ma, X.~Xiao, and Z.-X. Yang, ``Design and implementation of an
  adaptive neural network observer–based backstepping sliding mode controller
  for robot manipulators,'' \emph{Transactions of the Institute of Measurement
  and Control}, vol.~0, no.~0, p.~0, 2023.

\bibitem{xi2020adaptive}
R.~Xi, Z.~Yang, and X.~Xiao, ``Adaptive neural network observer based
  pid-backstepping terminal sliding mode control for robot manipulators,'' in
  \emph{2020 IEEE/ASME International Conference on Advanced Intelligent
  Mechatronics (AIM)}.\hskip 1em plus 0.5em minus 0.4em\relax IEEE, 2020, pp.
  209--214.

\bibitem{chen2015disturbance}
W.-H. Chen, J.~Yang, L.~Guo, and S.~Li, ``Disturbance-observer-based control
  and related methods—{A}n overview,'' \emph{IEEE Transactions on Industrial
  Electronics}, vol.~63, no.~2, pp. 1083--1095, 2015.

\bibitem{liang2022adaptive}
S.~Liang, R.~Xi, X.~Xiao, and Z.~Yang, ``Adaptive sliding mode disturbance
  observer and deep reinforcement learning based motion control for
  micropositioners,'' \emph{Micromachines}, vol.~13, no.~3, p. 458, 2022.

\bibitem{rabiee2019continuous}
H.~Rabiee, M.~Ataei, and M.~Ekramian, ``Continuous nonsingular terminal sliding
  mode control based on adaptive sliding mode disturbance observer for
  uncertain nonlinear systems,'' \emph{Automatica}, vol. 109, p. 108515, 2019.

\bibitem{zhu2018adaptive}
Y.~Zhu, J.~Qiao, and L.~Guo, ``Adaptive sliding mode disturbance observer-based
  composite control with prescribed performance of space manipulators for
  target capturing,'' \emph{IEEE Transactions on Industrial Electronics},
  vol.~66, no.~3, pp. 1973--1983, 2018.

\bibitem{shi2022adaptive}
D.~Shi, J.~Zhang, Z.~Sun, and Y.~Xia, ``Adaptive sliding mode disturbance
  observer-based composite trajectory tracking control for robot manipulator
  with prescribed performance,'' \emph{Nonlinear Dynamics}, vol. 109, no.~4,
  pp. 2693--2704, 2022.

\bibitem{xi2022adaptive}
R.-D. Xi, X.~Xiao, T.-N. Ma, and Z.-X. Yang, ``Adaptive sliding mode
  disturbance observer based robust control for robot manipulators towards
  assembly assistance,'' \emph{IEEE Robotics and Automation Letters}, vol.~7,
  no.~3, pp. 6139--6146, 2022.

\bibitem{bechlioulis2009adaptive}
C.~P. Bechlioulis and G.~A. Rovithakis, ``Adaptive control with guaranteed
  transient and steady state tracking error bounds for strict feedback
  systems,'' \emph{Automatica}, vol.~45, no.~2, pp. 532--538, 2009.

\bibitem{cheng2022prescribed}
F.~Cheng, B.~Niu, L.~Zhang, and Z.~Chen, ``Prescribed performance-based
  low-computation adaptive tracking control for uncertain nonlinear systems
  with periodic disturbances,'' \emph{IEEE Transactions on Circuits and Systems
  II: Express Briefs}, vol.~69, no.~11, pp. 4414--4418, 2022.

\bibitem{wang2021event}
H.~Wang, M.~Li, C.~Zhang, and X.~Shao, ``Event-based prescribed performance
  control for dynamic positioning vessels,'' \emph{IEEE Transactions on
  Circuits and Systems II: Express Briefs}, vol.~68, no.~7, pp. 2548--2552,
  2021.

\bibitem{lu2009sliding}
Y.-S. Lu, ``Sliding-mode disturbance observer with switching-gain adaptation
  and its application to optical disk drives,'' \emph{IEEE Transactions on
  Industrial Electronics}, vol.~56, no.~9, pp. 3743--3750, 2009.

\end{thebibliography}

%








\end{document}